\documentclass[journal]{IEEEtran}

\usepackage{graphicx}
\usepackage{xcolor}
\usepackage{url}
\usepackage{amsmath,amsfonts}
\usepackage{array}
\usepackage[utf8]{inputenc}

\usepackage{tikz}
\usetikzlibrary{positioning,calc}

\usepackage{cite}
\let\citep\cite
\let\citet\cite

\usepackage[caption=false,font=footnotesize]{subfig}

\DeclareUnicodeCharacter{FF0C}{,}

\usepackage{arydshln}
\usepackage{algorithm}
\usepackage{algpseudocode}
\usepackage{multirow}
\usepackage{stfloats}
\usepackage{enumitem}
\usepackage{verbatim}
\usepackage{balance}  

\begin{document}

\title{Batch Normalization–Free Fully Integer Quantized Neural Networks via Progressive Tandem Learning}

\author{
Pengfei Sun, Wenyu Jiang$^*$, ~\IEEEmembership{Senior Member, IEEE}, Piew Yoong Chee, Paul Devos, and Dick Botteldooren,~\IEEEmembership{Senior Member, IEEE}, 
\thanks{Pengfei Sun, Paul Devos, and Dick Botteldooren are with the Department of Information Technology, WAVES Research Group, Ghent University,Ghent, Belgium.}
\thanks{Wenyu Jiang is and Piew Yoong Chee was with Institute for Infocomm Research (I$^2$R), Agency for Science, Technology and Research (A*STAR), Singapore ($^{*}$Corresponding Author${:}$ wjiang@i2r.a-star.edu.sg)
}}

\maketitle

\begin{abstract}
Quantised neural networks (QNNs) shrink models and reduce inference energy through low-bit arithmetic, yet most still depend on a running statistics batch normalisation (BN) layer, preventing true integer-only deployment. Prior attempts remove BN by parameter folding or tailored initialisation; while helpful, they rarely recover BN’s stability and accuracy and often impose bespoke constraints. We present a BN-free, fully integer QNN trained via a progressive, layer-wise distillation scheme that slots into existing low-bit pipelines. Starting from a pretrained BN-enabled teacher, we use layer-wise targets and progressive compensation to train a student that performs inference exclusively with integer arithmetic and contains no BN operations. On ImageNet with AlexNet, the BN-free model attains competitive Top-1 accuracy under aggressive quantisation. The procedure integrates directly with standard quantisation workflows, enabling end-to-end integer-only inference for resource-constrained settings such as edge and embedded devices.
\end{abstract}


\section{Introduction}

Deep neural network quantization has become an indispensable approach for reducing both memory footprint and computational complexity, thereby enabling deployment of large models on resource‐constrained hardware \citep{han2015deep,jacob2018quantization,wang2023toward}. By representing weights and activations in low‐bit formats (e.g., 8‐bit or even binary), quantized networks can dramatically reduce energy consumption and on‐chip storage while fully leveraging integer‐only accelerators \citep{courbariaux2016binarized,esser2019learned}. Various quantization schemes—uniform or non-uniform fixed-point, mixed-precision, binary networks, and teacher-student distillation—have been proposed to trade off model size, inference speed, and accuracy \citep{choi2018pact, LEROUX2020534, wang2019haq}.

Despite these advances, most quantized training and inference pipelines still rely on Batch Normalization (BN) layers to stabilize activations and accelerate convergence \citep{ioffe2015batch,courbariaux2016binarized}. However, BN depends on full‐precision mean and variance statistics \citep{ioffe2015batch}, and its affine transformations are typically folded into high‐precision parameters at inference \citep{jacob2018quantization}. This reliance disrupts the pure low‐bit computation graph, incurs extra memory accesses for statistics, and undermines end‐to‐end integer-only deployment on specialized hardware. Although methods such as folding BN parameters into weights \citep{jacob2018quantization} or redesigning weight distributions during training \citep{choi2018pact,wu2018training,yang2022towards,yang2020training} have been proposed, they often impose architectural constraints or fail to fully recover BN’s benefits under aggressive quantization. 
Moreover, current memristor-based analog computing architectures cannot natively support networks that employ batch normalization \citep{lyu2024memristive}.

To address these limitations, we introduce a progressive tandem‐learning framework that removes BN entirely while preserving the quantized accuracy and strategy of each low-precision network training. Leveraging layer‐wise knowledge distillation and a lightweight per-layer scale factor, our method distills a fully integer student from a pretrained BN-enabled quantized teacher using local loss functions. The resulting hardware‐friendly, BN‐free network is trained and deployed with pure integer arithmetic, achieving competitive ImageNet accuracy. Furthermore, our framework can be applied to any existing BN-based quantization algorithm: a BN-enabled quantized model can serve directly as teacher to produce a fully integer student in one training pass. 

\begin{itemize}
  \item We propose a Fully-Integer Quantized Neural Network (FIQNN) distilled from a pretrained, BN-enabled quantized model, eliminating the need for BN through a local progressive training scheme.
  \item Our method operates as a plug-in module without extra constraints on weight distributions or gradients, making it compatible with various quantized frameworks.
  \item We validate FIQNN on ImageNet, demonstrating competitive accuracy under aggressive quantization settings.
\end{itemize}

The remainder of this paper is organized as follows. Section II reviews related work. Section III describes the proposed progressive tandem-learning framework. Section IV presents experimental results and analysis. Section V concludes and outlines future work.

\section{Related Works}

\subsection{Quantized Neural Networks}
Quantized neural networks (QNNs) lower the precision of weights and activations to reduce bit-width, thereby decreasing model size, reducing memory bandwidth requirements, and lowering inference latency on specialized hardware. Early methods such as BinaryConnect and Binarized Neural Networks (BNNs) constrain weights and/or activations to \(\{\pm1\}\) using straight-through estimators for backpropagation \citep{courbariaux2015binaryconnect,courbariaux2016binarized}. Subsequent work extends this concept to multi-bit quantization: DoReFa-Net employs uniform \(m\)-bit quantization for weights, activations, and gradients \citep{b2}. Mixed-precision techniques assign bit-widths per layer to balance accuracy and hardware constraints \citep{wang2019haq,xu2023hybrid}. These advances enable QNNs to approach full-precision performance on large-scale benchmarks while operating exclusively in integer arithmetic.

\subsection{Knowledge Distillation}
Knowledge distillation (KD) transfers information from a high-capacity teacher model to a smaller or resource-constrained student by aligning output distributions or intermediate feature representations \citep{hinton2015distilling}. KD has been extended to heterogeneous tasks and cross-modal scenarios, such as transferring from visual or auditory models to EEG-based networks \citep{sun2025delayed}, speech enhancement \citep{han2024distil}, and from artificial neural networks to spiking neural networks \citep{wu2021tandem,yang2022training}. In quantization, KD mitigates accuracy loss due to low-precision representations. For example, Apprentice uses a full-precision teacher to supervise a quantized student, yielding improvements on ImageNet \citep{mishra2018apprentice}. Quantization-aware distillation combines feature-level and logit-level losses to better preserve teacher knowledge under aggressive bit-width constraints \citep{polino2018model}. However, most low-bit networks rely on normalization layers—especially Batch Normalization—for training stability. Direct layer-wise distillation without BN may lead to error accumulation; progressive, layer-wise distillation strategies have been proposed to iteratively compensate for such errors \citep{wu2021progressive}. Our framework builds on these insights, applying local, layer-wise distillation losses to train a fully-integer student network without any BatchNorm layers.

\section{Proposed Methods}

This section describes our proposed methodology, which comprises progressive knowledge distillation for training fully-integer neural networks. To establish a solid foundation, we first revisit quantized neural networks (QNNs) before detailing our approach.

\subsection{Algorithm for Training Low-Precision Weights and Activations in Teacher Models}

To validate our framework, we adopt the DoReFa-Net quantization scheme \citep{b2}. In this paradigm, a real-valued input \(r_i\in[0,1]\) is quantized to an \(m\)-bit integer \(r_o\) via
\begin{equation}
r_o = \mathrm{round}\bigl((2^{m}-1)\,r_i\bigr),
\end{equation}
where \(\mathrm{round}(\cdot)\) denotes the nearest-integer mapping and \(r_o\in[0,2^{m}-1]\). The integer \(r_o\) is then rescaled to match the desired range. For activations (in \([0,1]\)), we apply
\begin{equation}
r_o \leftarrow \frac{r_o}{2^{m}-1}\,.
\end{equation}
For weights (in \([-1,1]\)), we instead use
\begin{equation}
r_o \leftarrow \frac{2\,r_o}{2^{m}-1} \;-\; 1.
\end{equation}
Although gradients may also be quantized similarly, our primary emphasis is on inference efficiency, so we allow gradients to remain in full precision.

\begin{figure*}[!t]
    \centering
    \subfloat[Stage 1: Initialization of quantized weights and per-layer scale factors using the teacher’s local activations.\label{fig:kdqnn_stage1}]{
      \includegraphics[width=0.90\linewidth]{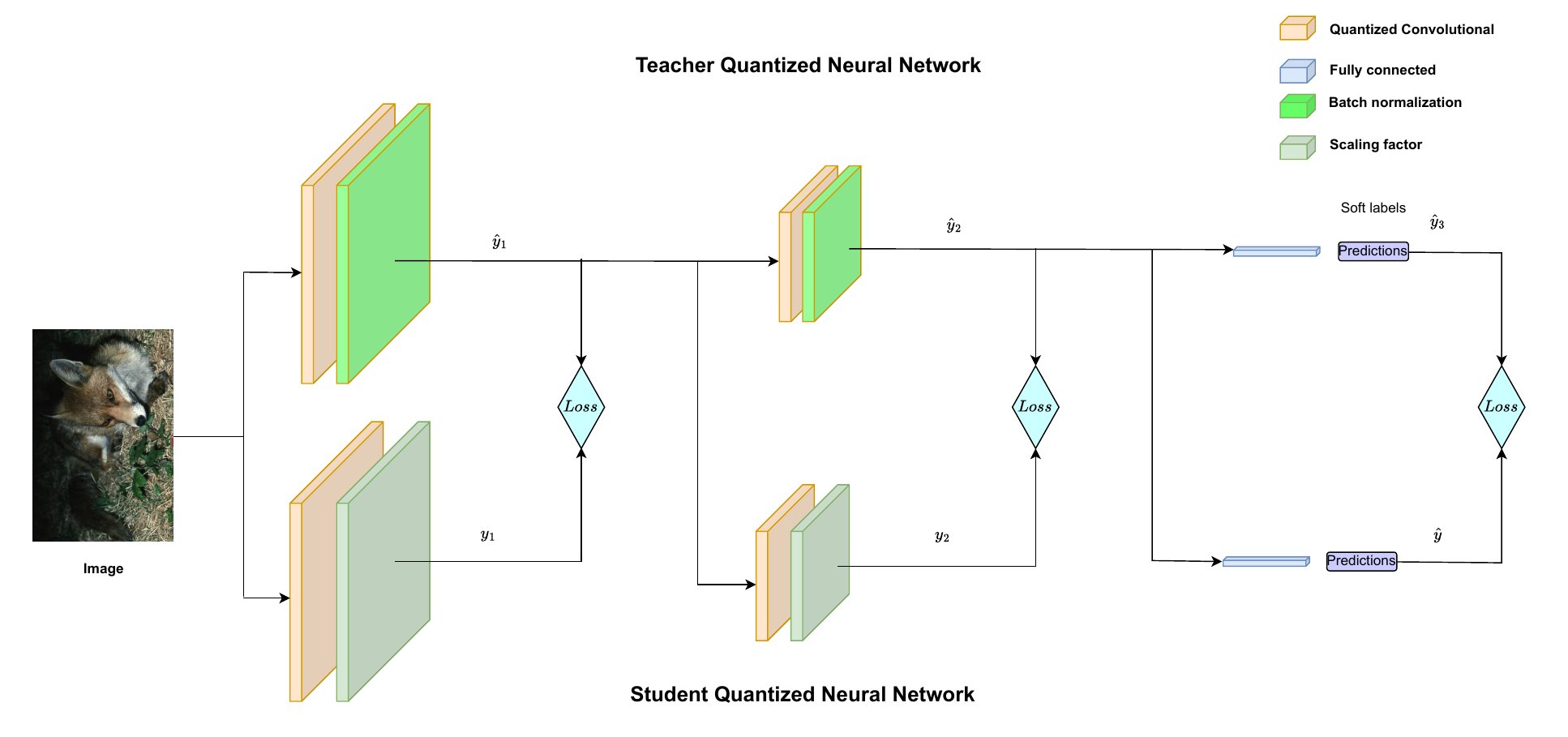}
    }\\
    \subfloat[Stage 2: Layer-wise progressive tandem learning, where all preceding layers remain frozen and only the current layer is optimized.\label{fig:kdqnn_stage2}]{
      \includegraphics[width=0.35\linewidth]{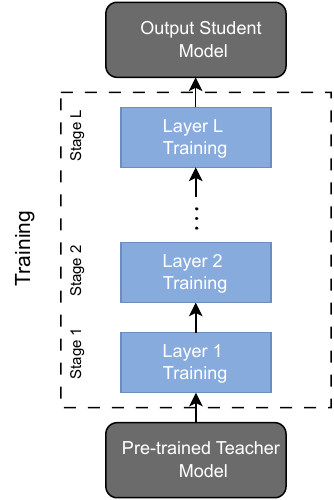}
    }\hfill
    \subfloat[Stage 3: Detailed view of quantized weight initialization guided by the teacher’s activations.\label{fig:kdqnn_stage3}]{
      \includegraphics[width=0.55\linewidth]{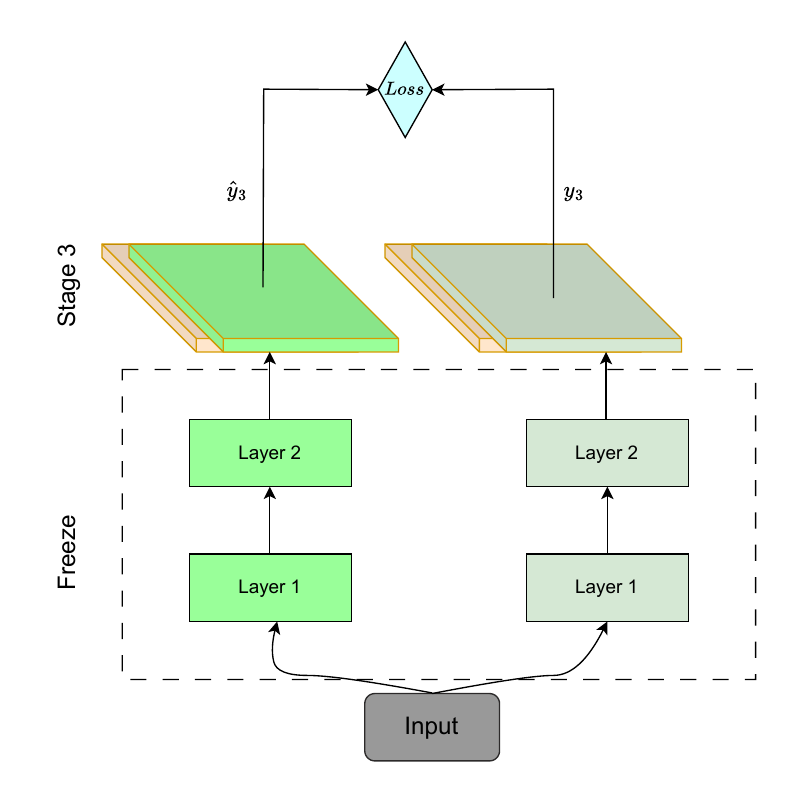}
    }
    \caption{ Overview of the FIQNN training process.}
    \label{fig:kdqnn}
\end{figure*}

\subsection{Progressive QNN-to-FIQNN Knowledge Distillation Method}

Our aim is to eliminate BN and keep inference strictly integer-only. To retain the benefits of existing quantisation methods while achieving competitive accuracy without BN, we adopt a tandem, block-wise scheme: a fixed full-precision-with-BN teacher supervises only the current block, while the student for that block is BN-free and integer; the remaining blocks stay full precision until their turn. In parallel, we introduce a fixed, integer-friendly scale re-balancing that substitutes for BN’s stabilising effect during training. This localises and corrects mismatches within each block, preventing scale drift and activation saturation from accumulating. Building on these ideas, we use a progressive distillation framework to transfer knowledge from the BN-enabled teacher to the BN-free, fully integer student \citep{gou2021knowledge}. Figure~\ref{fig:kdqnn} summarises the three-stage process.

\subsubsection{Teacher Model and Student Pre-Training}

We begin with a pretrained QNN teacher that employs \(W\)-bit weights and \(A\)-bit activations, including full-precision BN layers. The student shares the same bit-width configuration \((W,A)\), except it omits all BN layers. To compensate, we insert a lightweight, layer-wise scale factor immediately after each convolutional or fully connected operation in the student. This scale layer, parameterized by an integer factor \(\alpha\), approximates BN’s effect by re-centering and re-scaling activations so that their mean and variance approach 0 and 1, respectively. During pre-training, we freeze the teacher’s parameters and introduce a local loss for each student layer:
\begin{equation}
L = \bigl\lVert T(x_{\ell},\,\mathrm{BN}) \;-\;\alpha_{\ell}\,S(x_{\ell})\bigr\rVert_{2}^{2},
\end{equation}
where \(T(x_{\ell},\,\mathrm{BN})\) denotes the teacher’s output after BN at layer \(\ell\), and \(S(x_{\ell})\) is the student’s activation before scaling. Minimizing this loss jointly initializes both the quantized weights and the per-layer scale factor \(\alpha_{\ell}\). Once pre-trained, each \(\alpha_{\ell}\) is fixed for subsequent stages. During inference, the scale factor is incorporated into the quantization step as
\begin{equation}
r_{\!o} \leftarrow \frac{\alpha\,r_{\!o}}{2^{m}-1}\,,
\end{equation}
thus preserving pure-integer arithmetic throughout.

\subsubsection{Layer-Wise Progressive Tandem Training}

After pre-training, we perform layer-wise progressive tandem learning over \(L\) stages (where \(L\) is the total number of layers). In stage \(i\), we freeze all student layers \(\{1,\dots,i-1\}\) and train only layer \(i\). We compute a local distillation loss between the student’s output \(y_{i}\) at layer \(i\) and the teacher’s corresponding output \(\hat{y}_{i}\):
\begin{equation}
L_{i} \;=\; \bigl\lVert \hat{y}_{i} \;-\; y_{i}\bigr\rVert_{2}^{2},
\end{equation}
where the inputs to layers \(1\) through \(i-1\) are replaced with the teacher’s frozen activations. By minimizing \(L_{i}\), we optimize only the quantized weights in layer \(i\). Figure~\ref{fig:kdqnn_stage2} illustrates this process. Once layer \(i\) converges, its parameters are frozen, and we proceed to stage \(i+1\) until all layers are trained (see Figure~\ref{fig:kdqnn_stage3} for initialization details).

\subsubsection{Fully Trained Integer-Based Network}

Upon completing all \(L\) stages, the student model is fully trained and contains no BN layers—only quantized weights and fixed per-layer scale factors. At inference time, each layer simply performs integer-valued convolutions or matrix multiplications; no floating-point computations or dynamic normalization operations are needed. As a consequence, the model executes exclusively with integer arithmetic, reducing both computational complexity and memory footprint. This streamlined design makes the network especially well suited for deployment on resource-constrained hardware, where low-latency, low-power inference is critical.

\begin{algorithm}[!t]
\caption{Progressive Tandem Learning for BN-Free Fully-Integer QNN}
\label{alg:fiqnn-brief}
\textbf{Input:} Pretrained quantized teacher $\mathcal{T}$ (with BN), training data $\mathcal{D}$ \\
\textbf{Output:} BN-free, fully-integer student $\mathcal{S}$
\begin{algorithmic}[1]
\State \textbf{Stage 1 (Fig.~\ref{fig:kdqnn_stage1}): Local init}
\For{$\ell=1\ \to\ L$}
  \State Initialise $\mathcal{S}$ layer-$\ell$ (quantized weights)
  \State Learn a fixed integer-friendly scale $\alpha_\ell$ by matching teacher post-BN activations
  \State Freeze $\alpha_\ell$ (fold into the quantization path)
\EndFor

\State \textbf{Stage 2 (Fig.~\ref{fig:kdqnn_stage2} and Fig.~\ref{fig:kdqnn_stage3}): Progressive tandem}
\For{$i=1\ \to\ L$}
  \State Freeze student layers $1{:}i{-}1$
  \State Train only layer $i$ by aligning its output with teacher layer-$i$ target
  \State Freeze layer $i$
\EndFor
\State Run inference with integer transformation layer and fixed rescaling only

\State \textbf{return} $\mathcal{S}$
\end{algorithmic}
\end{algorithm}

\section{Experiments and Results}

\subsection{Implementation Details}
Our evaluation uses the ILSVRC-2012 (ImageNet) benchmark with AlexNet \citep{b3}, trained using the TensorPack configuration and hyperparameters \citep{tensorpack}. We adopt the DoReFa-Net low-precision training scheme throughout \citep{b2}. In addition, we assess performance on CIFAR-10 with a VGG-Small architecture \citep{simonyan2014very}.

\begin{table}[!t]
\centering
\caption{Top-1 and Top-5 error rates on ImageNet for various bit-width configurations under the DoReFa-Net framework, comparing BN-enabled teacher models to our BN-free student models. Baseline results are reproduced from DoReFa-Net \citep{b2}. }
\label{tbl:ablationstudy}
\begin{tabular}{cccccc}
\hline
Model & W & A  & Top-1 error & Top-5 error \\
\hline
FP32\citep{b2}                   & 32 & 32  & 44.10\% & –     \\ 
\hline
\multicolumn{6}{l}{\textbf{8-bit weights and activations}} \\ 
Teacher (BN-enabled)     & 8  & 8  & 47.00\% & 21.74\% \\
Student (BN-free)        & 8  & 8   & 47.72\% & 23.80\% \\ 
\hline
\multicolumn{6}{l}{\textbf{4-bit weights and activations}} \\ 
Teacher (BN-enabled)     & 4  & 4   & 48.50\% & 25.20\% \\
Student (BN-free)        & 4  & 4  & 49.45\% & 27.48\% \\ 
\hline
\end{tabular}
\end{table}

\begin{table}[!t]
\centering
\caption{CIFAR-10 accuracy (\%) on VGG-Small under 1-bit weights (W) and 1-bit activations (A). 
Both methods use the same training protocol and data preprocessing.}
\label{tab:cifar10_vggsmall_1bit}
\begin{tabular}{lccc}
\hline
\textbf{Method} & \textbf{W} & \textbf{A} & \textbf{Accuracy} \\
\hline
DoReFa-Net & 1 & 1 & 90.2\% \\
Ours (BN-free) & 1 & 1 & 90.8\% \\
\hline
\end{tabular}
\end{table}

\subsection{Results}
We evaluated our progressive tandem learning framework on ImageNet using AlexNet under two precision configurations: \((W,A,G)=(8,8,32)\) and \((4,4,32)\). Table~\ref{tbl:ablationstudy} summarizes Top-1 and Top-5 error rates for both the BN-enabled teacher models and our fully-integer, BN-free student models.

\begin{itemize}[leftmargin=*]
  \item \textbf{8-bit weights and activations.}  
    The BN-enabled teacher achieves a Top-1 error of 47.00\,\% and Top-5 error of 21.74\,\%.  
    After applying our progressive tandem learning (without any BN layers), the 8-bit student incurs only a minor drop to 47.72\,\% (Top-1) and 23.80\,\% (Top-5), indicating that our initialization and scale‐layer pre‐training recover most of the original BN benefits.
  \item \textbf{4-bit weights and activations.}  
    Under more aggressive quantization, the BN-enabled teacher records a Top-1 error of 48.50\,\% and Top-5 error of 25.20\,\%.  
    Our BN-free student sees a modest increase to 49.45\,\% (Top-1) and 27.48\,\% (Top-5).  
\end{itemize}

Our fully integer QNN, trained without any full-precision batch-normalisation layers, delivers competitive results on ImageNet. Even at 4-bit precision, the Top-1 error rises by no more than 1.0 percentage point compared with the BN-enabled teacher. This indicates that the layer-wise knowledge distillation and tandem training strategy retains accuracy while permitting end-to-end integer inference. We further assess generality on CIFAR-10 with VGG-Small in a stringent 1-bit/1-bit regime; under the same training protocol and preprocessing, the BN-free model matches the strong baseline (Table \ref{tab:cifar10_vggsmall_1bit}).

\section{Discussion and Conclusion}

We have presented a fully-integer quantization framework that eliminates all Batch Normalization layers by leveraging progressive, layer-wise knowledge distillation combined with lightweight integer scale layers. On the ImageNet classification task, our BN-free student achieves competitive results compared to the BN-based teacher model. This framework can plug into existing low-bit quantization pipelines with minimal modifications, yielding a hardware-friendly inference pipeline.

The progressive, layer-wise training schedule may increase wall-clock time compared to conventional quantization-aware training; however, these one-time costs are often offset by substantial runtime and energy savings during inference. Due to current resource limitations, we have evaluated our framework only on DoReFa-Net with two precision settings; future work should assess additional architectures and quantization configurations to further validate and generalize the approach and deploy on real hardware. Our current use of fixed, per-layer integer scales may demand careful calibration in very deep or highly nonlinear networks—preliminary evidence suggests that adaptive, learnable per-channel scales can better match activation distributions and leverage network heterogeneity \citep{sun2023adaptive,perez2021neural}. Finally, we will evaluate robustness under hardware-induced noise (e.g., analog inference, low-voltage operation) and combine our quantization strategy with cell-stitching techniques to bolster reliability on emerging accelerator platforms \citep{sun2024cell}. 
Our method is complementary to advanced model-compression strategies (e.g., pruning, low-rank factorisation), which we plan to explore to further reduce network complexity.

\balance
\begingroup
\small 
\clearpage
\bibliography{name}{}
\bibliographystyle{IEEEtran}
\endgroup

\end{document}